\documentclass[runningheads]{llncs}

\usepackage{eccv}

\usepackage{eccvabbrv}

\usepackage{graphicx}
\usepackage{booktabs}

\usepackage{multirow}
\usepackage[table]{xcolor}
\usepackage{adjustbox}
\usepackage{makecell}

\definecolor{rred}{RGB}{245, 152, 153}
\definecolor{oorange}{RGB}{253, 205, 154}
\definecolor{yyellow}{RGB}{248,244,140}

\usepackage[accsupp]{axessibility}  % Improves PDF readability for those with disabilities.

\usepackage{hyperref}

\usepackage{orcidlink}
\begin{document}
\title{CoIn: Comprehensive 2D-3D Inpainting with Gaussian Splatting Guidance}

\titlerunning{CoIn}

% % TODO FINAL: Replace with your author list. 
% % Include the authors' OCRID for the camera-ready version, if at all possible.
\author{Hana Kim\inst{1,2} \and
Minje Kim\inst{2} \and
Tae-Kyun Kim\inst{2}}

% % TODO FINAL: Replace with an abbreviated list of authors.
\authorrunning{H.~Kim et al.}

% % TODO FINAL: Replace with your institution list.
\institute{LG Electronics, Seoul, South Korea 
\email{hana1106.kim@lge.com} \and
KAIST, Daejeon, South Korea 
\email{\{minjekim, kimtaekyun\}@kaist.ac.kr}}

\maketitle

\begin{center}
\centering
\captionsetup{type=figure}
\includegraphics[width=\textwidth]{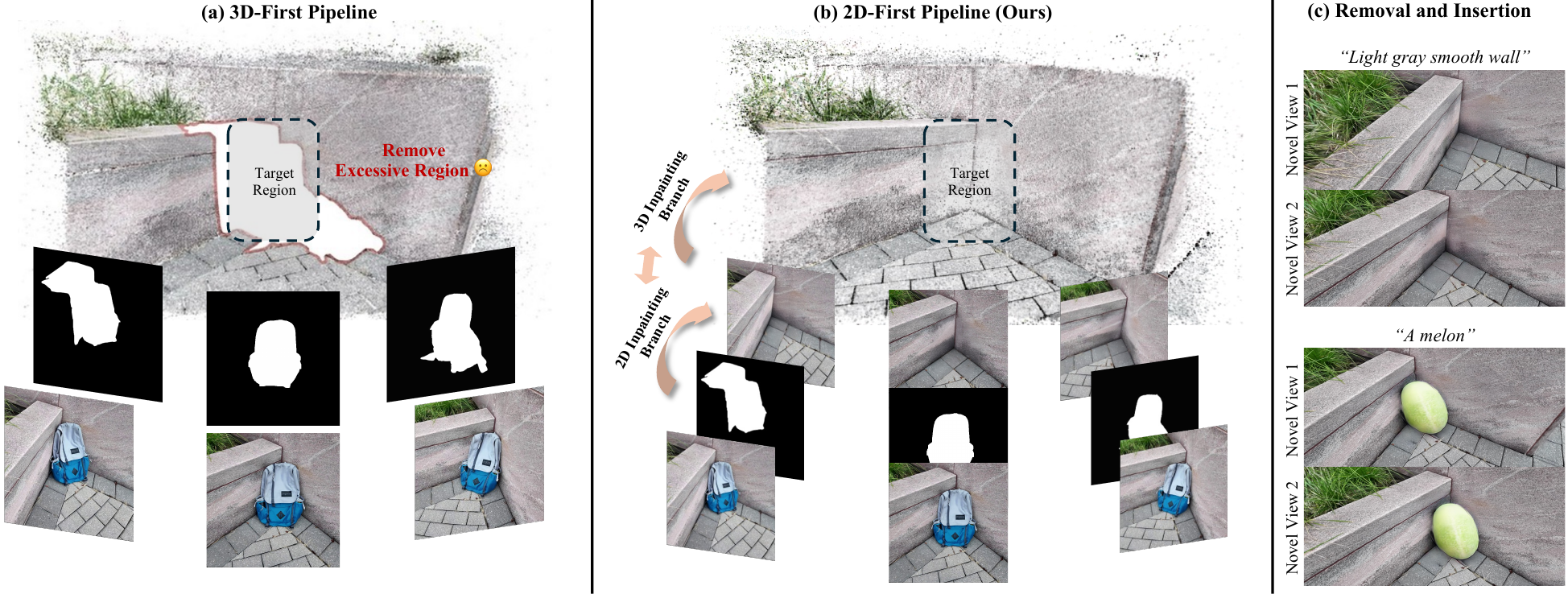}
\caption{
% \textbf{Comparing the impact of inconsistent 2D segmentation masks on 3D-first and 2D-first pipelines.} 
\textbf{Key impact of inconsistent masks on 3D-first vs. 2D-first pipelines. }
(a) Prior works, which typically adopt a 3D-first pipeline, often fail under inconsistent 2D segmentation masks across views, leading to excessive removal of regions and incorrect 3D segmentation. (b) In contrast, CoIn utilizes a 2D-first pipeline that comprehensively integrates 2D and 3D inpainting branches under Gaussian Splatting guidance. This approach ensures spatial and semantic consistency, enabling the use of arbitrary-shaped masks and supporting diverse tasks such as object removal and insertion (c).}
\label{Teaser}
\end{center}

\vspace{-0.3cm}
\begin{abstract}
  3D scene inpainting is essential for reconstructing areas corrupted by occlusions or limited viewpoints. While recent methods leverage Gaussian Splatting (GS) for efficient 3D editing, they often depend on precise multi-view segmentation masks and are inherently constrained to object removal tasks. We propose CoIn, a novel framework that bridges 2D inpainting models and 3DGS through a multi-stage consistency pipeline. Our approach first generates initial inpainted images using a diffusion model, enabling the use of arbitrary-shaped masks and diverse tasks like object insertion. We then introduce Reference Adaptive GS with Feature Attention to reconstruct a coarse 3D scene by adaptively weighing towards a reference view (2D $\rightarrow$ 3D). This 3D representation provides geometric guidance to the diffusion process via GS-based Reference Feature Warping, ensuring multi-view consistency (3D $\rightarrow$ 2D). Finally, a Texture-Enhancing Discriminator refines the 3D scene to achieve high photometric realism (2D $\rightarrow$ 3D). Experiments show that CoIn, effectively leveraging bidirectional information flow, achieves state-of-the-art performance and effectively handles both object removal and object insertion with flexible mask input.
  \vspace{-0.2cm}
\end{abstract}

\section{Introduction}
\label{sec:intro}

Inpainting 3D scene is essential for reconstructing incomplete or corrupted scenes arising from occlusions, sensor noise, or limited viewpoints. Recent advances in generative models have expanded this scope from simple object removal to editing and object insertion. To achieve this, various methods directly optimize Neural Radiance Fields (NeRF) \cite{mildenhall2020nerf} using priors such as RGB and depth \cite{mirzaei2023spin} or diffusion-based \cite{chen2024mvip}. However, the implicit representation of NeRF hinders direct geometry manipulation, necessitating complex volumetric optimization for precise removal, insertion, or shape editing.

To perform more geometric and efficient editing, subsequent methods \cite{wang2024learning, ye2024gaussian, shi2025imfine, huang20253d, wu2025aurafusion360} adopt the `3D-first' pipeline, which denotes the strategy of first reconstructing a 3D Gaussian Splatting (3DGS) \cite{kerbl20233d} scene from the original images to provide a structural basis before performing inpainting. However, this sequence requires accurate segmentation masks (e.g., obtained from SAM2 \cite{ravi2024sam}) across multiple views to precisely isolate target regions in 3D space. 
Furthermore, methods that initiate inpainting by pruning these reconstructed Gaussians, such as IMFine \cite{shi2025imfine}, 3DGIC \cite{huang20253d}, and AuraFusion \cite{wu2025aurafusion360}, are inherently restricted to removal tasks. Although GScream \cite{wang2024learning} avoids initial pruning, it is based only on a single 2D inpainted reference image, which limits its ability to maintain consistency across large viewpoint changes.

By contrast, the `2D-first' pipeline denotes a strategy that prioritizes performing 2D inpainting across multiple views, followed by leveraging 3D information, such as optical flow \cite{cao2024mvinpainter} or meshes \cite{barda2025instant3dit}, to ensure consistency among the generated images. Although pipelines like MVInpainter \cite{cao2024mvinpainter} or Instant3dit \cite{barda2025instant3dit} benefit from the diverse editing capabilities of 2D generative models, they often rely on fine-tuning with refined datasets or require additional accurate 3D labels.

% We propose a novel framework that integrates the versatile editing capabilities of 2D-first pipelines with the efficient optimization of 3DGS. By combining these two approaches, our method allows general mask shapes as input and performs diverse inpainting tasks with high efficiency. 
% Our 2D inpainter is guided by injecting 3D consistency loss based on Reference Adaptive Gaussian Splatting with Feature Attention, which adaptively controls the weight of multiple view information in 3D reconstruction. Moreover, we further improve the photometric realism of 3DGS renderings with our Texture Enhancing Discriminator.
% We demonstrate our method with both quantitative and qualitative experiments for object removal and insertion tasks.
We propose a novel framework that combines the flexibility of 2D-first pipelines with the efficient optimization of 3DGS. By integrating these approaches, our method supports arbitrary-shaped masks and handles diverse inpainting tasks with high efficiency. Multi-view consistency is maintained by guiding the 2D inpainter with 3D priors from Reference Adaptive GS, while a Texture Enhancing Discriminator further refines photometric realism. We demonstrate our approach through both quantitative and qualitative experiments for object removal and insertion tasks.
In summary, our contributions are as follows:
\begin{itemize}
\item We propose CoIn, a novel framework that seamlessly bridges generative 2D synthesis and explicit 3D reconstruction via a multi-stage pipeline.
\item We introduce Reference Adaptive Gaussian Splatting with Feature Attention (Ref-GS), which assigns adaptive weights to each view to optimize 3DGS toward the reference viewpoint.
\item We present Consistency Loss Guidance (CLG), which leverages GS-based Reference Feature Warping to enforce 3D consistency with the reference image during the denoising process. 
\item The Texture Enhancing Discriminator (TE-D) mitigates blurriness in generated patches by learning the distribution of real image patches. 
\end{itemize}

\section{Related Work}
\label{sec:related_work}

% 2.1.
\subsection{2D Image Inpainting}

2D inpainting refers to the task of reconstructing missing or masked regions in an image by leveraging contextual cues from visible areas.
Early approaches mainly filled missing areas by copying nearby content, often resulting in visually inconsistent completions \cite{efros1999texture}.
With the advent of neural networks trained on datasets, recent methods \cite{wang2024gridformer, lugmayr2022repaint, ju2024brushnet, saharia2022palette, wang2023towards} used combinations of convolution networks, such as GAN \cite{yu2022high}, Fourier convolution \cite{suvorov2022resolution}, or Wavelet decomposition \cite{jeevan2023wavepaint}.

Meanwhile, diffusion-based models are widely adopted for 2D inpainting due to their ability to produce diverse yet highly feasible results \cite{saharia2022palette, lugmayr2022repaint}. RePaint \cite{lugmayr2022repaint} extends the concept of unconditional DDPMs using the known region of an image as a condition, allowing the denoising process to serve as inpainting for the unknown region.
Recent approaches \cite{ju2024brushnet, wang2023towards} incorporate latent diffusion models \cite{rombach2022high} to inpainting tasks. Stable diffusion (SD) inpainting pipeline \cite{rombach2022high} takes the masked image as input and performs denoising directly in the latent space, achieving faster inference while producing plausible and realistic results.

However, using pure 2D image inpainters without any additional modification, stochastic sampling in the denoising process leads to severely inconsistent results given small viewpoint changes across multi-view images, even for the same images. These inconsistencies are not suitable for subsequent 3D reconstruction, causing inaccurate geometries and blurriness. 
While several methods \cite{deng2023mv, shi2023mvdream, poole2022dreamfusion, tang2024lgm, ai2024dream360, liu2023zero, shen2023anything, barda2025instant3dit, zhuang2024tip, cao2024mvinpainter, weber2024nerfiller, kim2025srhand} consider 3D consistency for 2D generation and editing, they often exhibit trade-offs between computational efficiency and structural flexibility. Specifically, MVInpainter \cite{cao2024mvinpainter} and NeRFiller \cite{weber2024nerfiller} attempt to enforce consistency via flow supervision or shared grid priors; however, these approaches still encounter limitations in mask shape flexibility, output resolution, or the need for per-dataset adaptation.

% 2.2.
\subsection{Guidance for Latent Diffusion Model}
There exist various strategies \cite{kim2024arbitrary, wang2025lldiffusion, ho2022classifier} to enable diffusion models to perform specific tasks or adapt to particular domains, specifically, fine-tuning \cite{hu2022lora}, incorporating a pretrained adapter \cite{ye2023ip, mou2024t2i}, or injecting guidance \cite{bansal2023universal, yu2023freedom, song2023loss} to align the model with a desired task or dataset. However, fine-tuning or training adapters generally relies on task-specific training data, which limits the flexibility of per-scene adaptation.

Instead, guidance-based control methods \cite{bansal2023universal, yu2023freedom, song2023loss} introduce additional control signals to the denoising process at inference time. FreeDoM \cite{yu2023freedom} formulates an energy function that measures the distance between the given condition $c$ and the noisy intermediate result $x_t$.
By minimizing the value of the energy function during denoising steps, diffusion models can generate desirable results without additional training.
Our method leverages guidance-based control to generate 3D-consistent inpainted images. We incorporate guidance from the trained 3DGS within the energy function for the inpainting diffusion model.

% 2.3.
\subsection{3D Scene Inpainting}
NeRF-based 3D inpainting methods \cite{mirzaei2023spin, chen2024mvip, lin2024taming} have recently shown strong performance by optimizing an implicit radiance field that can synthesize missing geometry and appearance from multi-view observations. However, the implicit formulation typically involves heavy computation and long training/rendering time, and explicit local edits often require re-optimization or complex volumetric procedures, limiting practical adaptation.

With 3D Gaussian Splatting (3DGS) \cite{kerbl20233d}, many subsequent approaches have performed inpainting explicitly in 3D space using point-based scene representations \cite{ye2024gaussian, shi2025imfine, wang2024learning, huang20253d, wu2025aurafusion360}. They usually get one reference image as a guide for inpainting the 3D scene. However, GScream \cite{wang2024learning} relies only on a single reference view that does not cooperate with other viewpoint images, large deviations from the reference view cause 3D inpainting to fail. 
In contrast, 3D-first pipelines remove objects directly in 3D and complete geometry using cues such as Laplacian smoothing to warp the reference image, thereby handling large viewpoint changes \cite{shi2025imfine, huang20253d, wu2025aurafusion360}. Although efficient, they rely on highly accurate 2D segmentation masks to specify a target object consistently across dense views. As shown in \cref{Teaser}, in the absence of 3D consistency in the masks, the segmented regions do not align in the 3D space. This leads to geometrically inconsistent completion or unintended deletion of non-target regions and degrades 3D stability. Moreover, these pipelines are designed primarily for object removal rather than general inpainting tasks, such as insertion.

To the best of our knowledge, CoIn represents a comprehensive framework that establishes a correlated inpainting pipeline by integrating the generative capabilities of 2D models with the explicit representation of 3D Gaussian Splatting. Our framework, CoIn, is designed to leverage both 2D and 3D strengths: a 2D inpainting branch enables semantically meaningful edits under arbitrary-shaped masks, while an explicit 3D inpainting branch ensures strict multi-view consistency. Unlike 3D-first pipelines, CoIn does not require accurate segmentation masks and supports both removal and insertion, while avoiding the cross-view inconsistency issues common in 2D multi-view inpainting.

\section{Preliminary}
\label{sec:preliminary}

\paragraph{3D Gaussian Splatting.} 3D Gaussian Splatting (3DGS) \cite{kerbl20233d} is an explicit point-based scene representation in which each Gaussian primitive is parameterized by its center position $\mu$, rotation matrix $R$, scale $S$, color $c$ and opacity $\alpha$.
The covariance matrix of each Gaussian is defined as
$\Sigma = R SS^T R^{\top}$, and Gaussians are represented by

\begin{equation}
\mathcal{G}(x) = \exp\!\left(-\tfrac{1}{2}(x - \mu)^{\top}\Sigma^{-1}(x - \mu)\right)
\end{equation}

Training is conducted by minimizing a combination of $\mathcal{L}_1$ and $\mathcal{L}_{SSIM}$ between the rendered image $R_n$ and the ground-truth RGB image $I_n$:
\begin{equation}
\mathcal{L}_R(R_n, I_n) =
(1-\lambda) \mathcal{L}_1 +
\lambda (1 - \mathcal{L}_{SSIM})
\end{equation}

\paragraph{Scaffold-GS.} Unlike the original 3DGS, which directly optimizes the parameters of distributed gaussians, Scaffold-GS \cite{lu2024scaffold} adopts a lightweight anchor-based structure. Each anchor has a learnable feature embedding (anchor feature), and compact MLPs decode this embedding into neural gaussian attributes within its voxel-defined local region. This hierarchical design reduces redundancy by performing densification at the anchor level, achieving comparable rendering quality to the original 3DGS. The sparse anchor structure also enables efficient point-based guidance to maintain geometric correspondence across views, while anchor features permit attention to be applied directly in 3D, thereby facilitating 3D inpainting. We adopt it as our base model for representing the 3D scene, and propose an efficient object handling solution.

 \begin{figure*}[t]
    \centering
    \includegraphics[width=\textwidth]{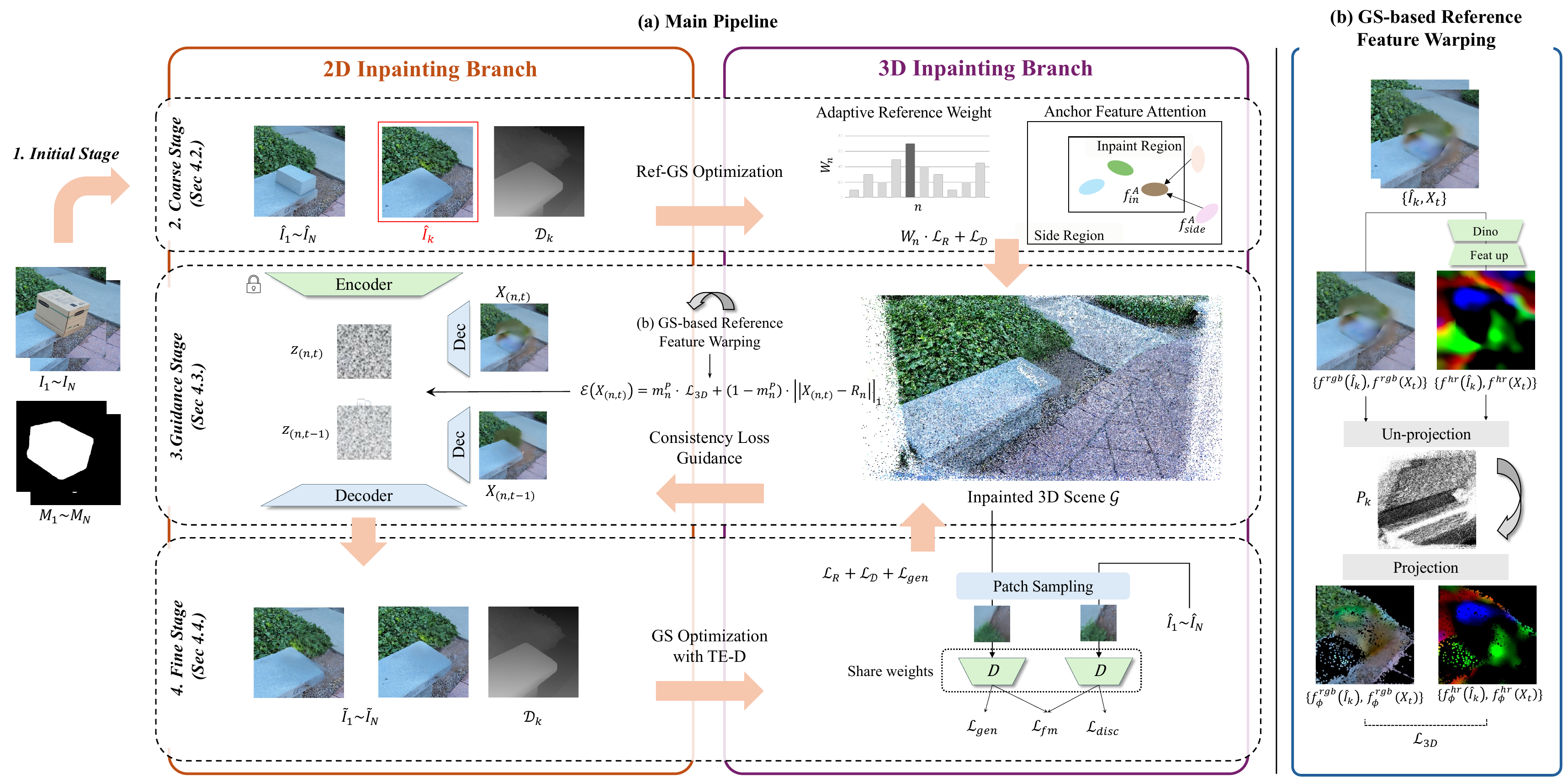}
    \caption{\textbf{Overview of CoIn}. We begin with initial 2D inpainting results and apply Reference GS with Feature Attention to obtain a coarse inpainted 3D scene. We then use Consistency Loss Guidance for the frozen latent diffusion inpainting model with GS-based Reference Feature Warping, and the consistency-preserved results are finally used to fine-tune the 3D Gaussian Splatting scene $\mathcal{G}$ with a Texture-Enhancing Discriminator for realistic details.}
    \label{fig:main_arch}
\end{figure*}

\section{Method}
\label{sec:method}

% 4.1
\subsection{Coherent 2D-3D Inpainting} 
\label{sec:method:overview}
\paragraph{Overview.} Our method aims to achieve 3D-consistent inpainting from a set of object-present input images $\{I_n\}_{n=1}^N $ and their corresponding masks $\{M_n\}_{n=1}^N$. We begin with initial inpainting results $\{\hat{I}_n\}_{n=1}^N$ obtained from a SD inpainting model, which produces visually plausible completions for each image but fails to achieve consistency across views. To address this issue, we construct a 3D Gaussian scene from these initial results and refine it to enforce multi-view coherence while preserving fine details.

\cref{fig:main_arch}a illustrates our pipeline. We first apply a coarse stage, the Reference Adaptive GS with Feature Attention up-weights a selected reference view $\hat{I}_k$ and down-weights the others via per-view weights, while regularizing anchor features in the inpainting region by attending to neighboring anchors.
It then enforces multi-view geometric and appearance consistency by warping reference features on the GS-derived point cloud and injecting them as guidance into the latent diffusion inpainting model. With the consistency-guided inpainted images, we refine the GS scene for high-frequency textures and photometric realism via an adversarial patch discriminator in a fine stage.

By integrating these components, our framework produces a 3D inpainted scene that is geometrically consistent, visually coherent, and seamlessly completed across all views.
We introduce the Reference Adaptive GS with Feature Attention (Ref-GS) in \cref{sec:method:RGO}, Consistency Loss Guidance (CLG) in \cref{sec:method:CLG}, and Texture Enhancing Discriminator (TE-D) in \cref{sec:method:TED}.

% 4.2
\subsection{Reference Adaptive GS with Feature Attention}
\label{sec:method:RGO}

We first construct a 3D Gaussian Splatting (3DGS) scene to provide 3D guidance to the 2D inpainting branch. However, employing vanilla GS often yields view-dependent inconsistencies and overly blurred renders, which are not suitable for multi-view inpainting. To stabilize the scene, we applied a per-view weight to the rendering loss, encouraging the GS scene to align with the reference view $\hat{I}_k$ while suppressing views that are inconsistent. Before optimization, we prune points whose projections fall inside the masks, thereby eliminating the influence of the residual object from the COLMAP initialization \cite{schonberger2016structure}.

For each view $n$, we compute a weight $W_n$ that up-weights the reference view $\hat{I}_k$ and down-weights views inconsistent with the reference view:

\begin{equation}
W_n =
\begin{cases}
\lambda_r, & \text{if } n = k \\
\dfrac{1}{1 + \exp(\lambda_r (\mathcal{L}_R^{(n,t)} - \mathcal{L}_R^{(n,t-1)}))}, & \text{if } n \neq k
\end{cases}
\end{equation}
\label{equation: refer_W}
where $\mathcal{L}_R^{(n,t)}$ represents the current photometric loss of the $n$-th view of $t$-th iteration, $\mathcal{L}_R^{(n,t-1)}$ is from the previous iteration of the same view, and $\lambda_r$ is a weight parameter to the reference view. This fuction leads the GS scene to better reflect the reference image.

Although GScream \cite{wang2024learning} employs cross-attention to propagate texture from the surrounding region into the inpainted region, it is sensitive to variations in view range since it is applied only to the reference view. To address this limitation, we use unidirectional attention on anchor features for every training view. We update the features $f^A_{\text{in}}$ via $\tilde {f^A_{\text{in}}} =\mathrm{Attn}(Q=f^A_{\text{in}},\ K=f^A_{\text{side}},\ V=f^A_{\text{side}})$ where $f^A_{\text{in}}$ and $f^A_{\text{side}}$ denote anchor features inside the inpainting region and in its neighborhood, respectively.
This update enhances the consistency between the inpainted region and the surrounding areas, resulting in more coherent textures.

%% original version
% Let $f^A_{\text{in}}$ and $f^A_{\text{side}}$ denote anchor features inside the inpainting region and in its neighborhood, respectively.
% We update the features $f^A_{\text{in}}$ via $\tilde {f^A_{\text{in}}} =\mathrm{Attn}(Q=f^A_{\text{in}},\ K=f^A_{\text{side}},\ V=f^A_{\text{side}})$, 
% thereby inheriting appearance cues from surrounding anchors and reducing artifacts when lifting 2D edits to 3D.

In addition, we apply a depth loss in the reference view using the estimated depth, making the GS-derived point cloud $P_k$ more stable and better aligned with the reference view geometry. We begin by using a monocular depth estimator \cite{yang2024depth} to estimate depth $\mathcal{D}_k$ from $\hat{I}_k$. With a linear transformation, we transform the rendering depth $R^\mathcal{D}$ to $\bar{R^\mathcal{D}} = A*R^\mathcal{D}+B$. A and B are obtained by solving a least-squares problem with respect to $\mathcal{D}_k$ \cite{wang2024learning, ke2024repurposing, yu2022monosdf}.
Then the total GS optimization loss is computed as follows:

\begin{equation}
    \mathcal{L} = W_n \cdot \mathcal{L}_R(R_n, \hat{I_n}) + \lambda_\mathcal{D}\mathcal{L}_\mathcal{D}
\end{equation}
\begin{equation}
    \mathcal{L}_\mathcal{D} = \frac{1}{HW}\sum||\bar{R^\mathcal{D}}-\mathcal{D}_k||_1
\end{equation}
where $\lambda_\mathcal{D}$ denotes the weight for the depth loss, and H and W are height and width of the image, respectively.
% This stage yields a GS scene that is both reference-aligned and feature-regularized, forming a stable basis for subsequent consistency guidance.

% 4.3
\subsection{Consistency Loss Guidance}
\label{sec:method:CLG}
We enforce cross‐view consistency among inpainted images with an energy-based guidance inspired by FreeDoM \cite{yu2023freedom}. During diffusion denoising at step $t$, we update the sample to minimize an energy function built from the GS scene (\cref{sec:method:RGO}), leveraging both its renderings and the reconstructed point cloud $P_k$. We improve the consistency between the reference image $\hat I_k$ and the current view decoded image $X_{(n,t)}= Dec(z_{(n,t)})$ from the denoising diffusion latent $z_{(n,t)}$ by introducing GS-based Reference Feature Warping (\cref{fig:main_arch}b). 
Given the point cloud $P_k$ and the camera parameters, we unproject per-view features into 3D and reproject them into an intermediate viewpoint $\phi$.  This viewpoint is established by interpolating between the reference ($\mathbf{P}_k$) and current ($\mathbf{P}_n$) camera poses; specifically, we apply linear interpolation for translation, $\mathbf{t}_\phi = (1-\alpha)\mathbf{t}_k + \alpha\mathbf{t}_n$, and Spherical Linear Interpolation (Slerp) for rotation, $\mathbf{q}_\phi = \text{Slerp}(\mathbf{q}_k, \mathbf{q}_n; \alpha)$, with $\alpha=0.5$. From each image, we extract (i) high-frequency features with DINO \cite{caron2021emerging} upscaled via Feat-Up \cite{fu2024featup}, and (ii) appearance (RGB) features.
% Following Met3R \cite{asim2025met3r},
Let $f_\phi(\cdot;P_k)$ denote the warp operator that projects the features to the intermediate view $\phi$ with $P_k$. We then form a 3D consistency loss by applying a cosine similarity term to the high-frequency features and a $L2$ similarity term to the RGB features:
\begin{equation}
    \mathcal{L}_{3D} = 1-cos(f_\phi^{hr}(\hat{I_k}), f_\phi^{hr}(X_{(n,t)})) + ||f_\phi^{rgb}(\hat{I_k})-f_\phi^{rgb}(X_{(n,t)})||_2^2
\end{equation}

% This GS-based Reference Feature Warping provides geometry-aware guidance to the diffusion at step t, enforcing cross-view consistency in both structure (via high-frequency features) and appearance (via RGB). 
% Due to imperfect $P_k$, we further use an $L1$ loss with GS rendering $R_n$.
% Therefore, the final energy function is 
% \begin{equation}
%     \mathcal E(X_{(n,t)}) = m_n^P \cdot \mathcal{L}_{3D} + (1-m_n^P) \cdot ||X_{(n,t)}-R_n||_1
% \end{equation}
% \label{equation: energy_f}
% \noindent with $n$-th view's point-cloud support mask $m^P_n$. Guidance is applied by decreasing $\mathcal E(X_{(n,t)})$, encouraging multi-view consistency and geometry-aware inpainting by leveraging correspondences from the GS scene within the mask.
This GS-based Reference Feature Warping provides geometry-aware guidance at step $t$ to enforce cross-view consistency. To specifically account for non-overlapping regions between the current view and the reference view, we introduce the point-cloud support mask $m_n^P$, which identifies areas where valid 3D correspondences are available. Furthermore, to compensate for imperfections in the reconstructed point cloud $P_k$, we incorporate an $L1$ loss against the GS rendering $R_n$ as a complementary guidance.

Consequently, the final energy function is formulated as:
\begin{equation}
    \mathcal E(X_{(n,t)}) = m_n^P \cdot \mathcal{L}_{3D} + (1-m_n^P) \cdot ||X_{(n,t)}-R_n||_1
\end{equation}
\label{equation: energy_f}
By minimizing the value of $\mathcal E(X_{(n,t)})$, our framework encourages multi-view consistency and geometry-aware inpainting by leveraging correspondences from the GS scene within the mask.

% 4.4
\subsection{Texture Enhancing Discriminator}
\label{sec:method:TED}

Although \cref{sec:method:CLG} produces per-view inpainting results, they still exhibit residual artifacts and blurriness due to imperfect geometry and strong guidance. We therefore introduce a Texture-Enhancing Discriminator (TE-D) that transfers only texture information from the initial inpainting output $\hat I_n$ to the GS renderings $R_n$. 
Specifically, TE-D is trained jointly with the GS model on small image patches, treating patches from $\hat I_n$ as real and patches from the current GS renderings as fake. The GS model (generator) is trained adversarially against the discriminator, enhancing high-frequency texture details while preserving the overall geometric structure. Operating in local patches \cite{isola2017image} near the inpainted region, it emphasizes texture fidelity rather than geometry, helping remove guidance-induced blur (see \cref{fig:full_abl}A(c)).
Simultaneously, the depth loss and the rendering loss used in the GS optimization are computed with the CLG inpainting images $\tilde I_n$.
\begin{equation}
    \mathcal{L} =\mathcal{L}_R(R_n, \tilde{I}_n) + \lambda_D \mathcal{L}_\mathcal{D} + \lambda_{gen}\mathcal{L}_{gen}(R_n, \hat{I}_n)
\end{equation}
\label{equation: loss with D}
where $\lambda_{gen}$ denotes the weight for the adversarial loss. The combination of adversarial texture matching to $\hat I_n$ and photometric agreement with $\tilde I_n$ fine-tunes the inpainted 3D scene $\mathcal{G}$, yielding high fidelity while maintaining the multi-view consistency established in the previous stage.

\section{Experiments}
\label{sec:experiments}

\subsection{Experimental Setup}
\paragraph{Datasets.} Following prior works, we perform experiments on the SPIn-NeRF dataset \cite{mirzaei2023spin}. The dataset contains 10 scenes, each providing 100 calibrated images, 60 object-present and 40 object-absent. For evaluation, we inpaint the object-present images using the provided segmentation masks or bounding boxes derived from them. After inpainting, we synthesize novel views at the object-absent viewpoints to compute the metrics. We also evaluated on the IMFine dataset \cite{shi2025imfine}, which comprises 20 scenes grouped into four coverage cases (90°, 120°, 180°, 360°) having broader view changes than SPIn-NeRF. Each scene contains 125 object-present and 75 object-absent images, along with binary masks and calibrated camera parameters. To better show the effect of the reference image, we evaluated ten scenes with $90\,^{\circ}$, $120\,^{\circ}$and $180\,^{\circ}$viewpoint coverage. The $360\,^{\circ}$scenes are shown in the supplementary materials.

\paragraph{Evaluation Metrics.} We report LPIPS \cite{zhang2018unreasonable}, PSNR, and FID \cite{heusel2017gans} on the full images. To show the performance fidelity in masked regions more effectively, we also report the masked variants, m-LPIPS, m-PSNR, and m-FID, computed only within the ground-truth segmentation mask. 
For the object insertion task, we follow established practices by calculating the $CLIP_{dir}$ (CLIP Text-Image Directional Similarity) to assess how well the generated 3D content aligns with the provided text instructions. Additionally, we conduct a user study to evaluate and compare our method with state-of-the-art baselines. Specifically, we utilize a comparative voting process to assess four key dimensions: multi-view consistency, overall visual quality, alignment with text prompts and absence of visual artifacts. Detailed experimental setups and the specific questions used in the user study are provided in the supplementary material.

\begin{table*}[t]
\centering
\caption{\textbf{Quantitative evaluation on the SPIn-NeRF dataset \cite{mirzaei2023spin}.} Note that m-LPIPS \cite{zhang2018unreasonable} and m-FID \cite{heusel2017gans} represent the LPIPS and FID scores within the ground truth segmentation masks. BB mask denotes a bounding-box mask derived from the segmentation mask. The top three results are highlighted in \textcolor{rred}{red}, \textcolor{oorange}{orange}, and \textcolor{yyellow}{yellow}, respectively.}
\begin{adjustbox}{scale=0.85}
\begin{tabular}{@{}c|c|c|cccc@{}}
\toprule
3D Representation                   & Mask Type                 &                   & m-LPIPS ($\downarrow$) & LPIPS ($\downarrow$) & m-FID ($\downarrow$) & FID ($\downarrow$) \\ \midrule
\multirow{3}{*}{NeRF}               & \multirow{7}{*}{Seg mask} & SPIn-NeRF  \cite{mirzaei2023spin}     & 0.053   & 0.31  & 153.4 & 49.6 \\
                                    &                           & MVIP-NeRF    \cite{chen2024mvip}     & 0.050   & 0.31  & 173.4 & 50.5 \\
                                    &                           & MALD-NeRF   \cite{lin2024taming}     & \cellcolor{oorange} 0.031   & 0.30  & 113.5 & 44.7 \\ \cmidrule(r){1-1} \cmidrule(l){3-7} 
\multirow{7}{*}{Gaussian Splatting} &                           & Gaussian Grouping \cite{ye2024gaussian} & 0.037   & \cellcolor{yyellow} 0.26  & 132.5 & 44.9 \\
                                    &                           & 3DGIC   \cite{huang20253d}          & \cellcolor{rred} 0.028   & \cellcolor{yyellow} 0.26  & 96.3  & 36.4 \\
                                    &                           & GScream  \cite{wang2024learning}         & \cellcolor{yyellow} 0.032   & \cellcolor{yyellow} 0.26  & \cellcolor{yyellow} 86.1  & \cellcolor{yyellow} 31.2 \\
                                    &                           & Ours              & \cellcolor{yyellow} 0.032   & \cellcolor{rred} 0.23  & \cellcolor{rred} 80.4  & \cellcolor{oorange} 28.9 \\ \cmidrule(l){2-7} 
                                    & \multirow{3}{*}{BB mask}  & 3DGIC             & 0.047   & 0.38  & 168.0 & 104.1 \\
                                    &                           & GScream           & 0.043   & 0.29  & 104.6 & 34.2 \\
                                    &                           & Ours              & 0.033   & \cellcolor{oorange} 0.24  & \cellcolor{oorange} 95.1  & \cellcolor{rred} 26.6 \\ \bottomrule
\end{tabular}
\end{adjustbox}
\label{tab:spinnerf_quantitative}
\end{table*}

\begin{figure*}[t]
    \centering
    \includegraphics[width=\textwidth]{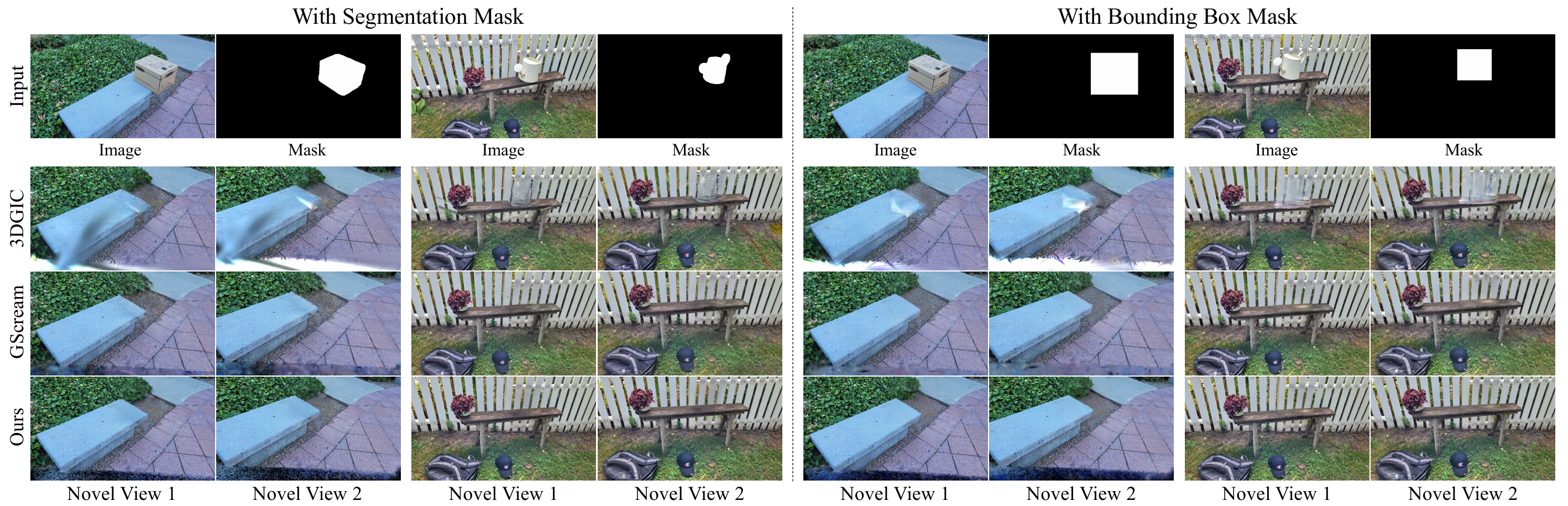}
    \caption{\textbf{Qualitative results on the SPIn-NeRF dataset \cite{mirzaei2023spin}}. The first row shows the input images and corresponding masks for each scene, followed by the inpainting results of the baseline methods 3DGIC \cite{huang20253d}, GScream \cite{wang2024learning}, and Ours in subsequent rows. By comparing the four columns on the left with the four columns on the right, we can observe the difference between using the segmentation masks and the bounding-box masks for inpainting.}
    \label{fig:qual_spin}
\end{figure*}

\paragraph{Implementation Details.} All experiments were conducted on a single NVIDIA RTX 4090 (24GB) GPU. We utilize Stable Diffusion 2.0 \cite{rombach2022high} for 2D inpainting, processing $512 \times 512$ mask-centered crops. The regions outside the mask are combined with the ground-truth image to preserve originality when training the GS model. Following prior works \cite{wang2024learning, shi2025imfine}, we empirically select a single reference view $\hat{I}_k$ from the initial 2D inpainting results to serve as a basis for maintaining consistency across the remaining views. For memory efficiency during the Consistency Loss Guidance (CLG) stage, we employ normalized DINO (dino16) \cite{caron2021emerging} and FeatUp \cite{fu2024featup} features extracted from $256 \times 256$ resized images. The TE-D module is trained in two distinct stages: first, we perform 10k iterations of fine-tuning the generator $\mathcal{G}$ on the CLG-inpainted images $\tilde{I}_n$, followed by 10k iterations of joint adversarial training. During this training, $64 \times 64$ patches are extracted via mask-based probabilistic sampling to focus on the inpainted regions. Detailed hyperparameter settings, including the loss weights $\lambda_D$, 
$\lambda_r$, and $\lambda_{gen}$ are provided in the supplementary material. The entire pipeline requires approximately 2.5 hours per scene, which is more efficient than typical learning-based 3D inpainting approaches.

\subsection{Quantitative Results}
\cref{tab:spinnerf_quantitative} presents quantitative results on the SPIn-NeRF dataset \cite{mirzaei2023spin} compared to baseline methods. For the setting using segmentation masks, we use the numbers reported in 3DGIC\cite{huang20253d} for SPIn-NeRF \cite{mirzaei2023spin}, MVIP-NeRF \cite{chen2024mvip}, MALD-NeRF \cite{lin2024taming}, Gaussian Grouping \cite{ye2024gaussian}, and 3DGIC \cite{huang20253d}. 
For fairness, we reproduce GScream \cite{wang2024learning} (with segmentation and bounding-box masks) and 3DGIC (with bounding-box masks) using official implementations. Since both GScream and 3DGIC require a single reference image, we follow the reference preparation protocol specified in each official implementation. Our method achieves state-of-the-art performance on most entries under both the segmentation and bounding-box settings. Notably, the performance gap between the two types of masks is small for our method, indicating that our method is less sensitive to mask shapes.

% \cref{tab:imfine_quantitative} compares our method with 3DGIC \cite{huang20253d}, GScream \cite{wang2024learning}, and IMFine \cite{shi2025imfine} on the IMFine dataset. Although IMFine reports metrics on cropped patches, we evaluate all models on full images for a unified comparison. For this purpose, we compute the metrics using the officially released result images for IMFine, while 3DGIC and GScream are evaluated through our reproduction using their official implementations. Under the segmentation mask setting, our method outperforms prior works across LPIPS \cite{zhang2018unreasonable}, PSNR, and FID \cite{heusel2017gans}.

\cref{tab:imfine_quantitative} compares our method with 3DGIC \cite{huang20253d}, GScream \cite{wang2024learning}, and IMFine \cite{shi2025imfine} on the IMFine dataset. For IMFine, we compute the metrics with officially released result images, while 3DGIC and GScream are evaluated through our reproduction using their official implementations. Our method outperforms baselines across LPIPS \cite{zhang2018unreasonable}, PSNR, and FID \cite{heusel2017gans} under the segmentation mask setting. 

%comparable scores to prior works on several metrics, while outperforming in terms of 

% Unlike these baselines, our model can handle arbitrary-shaped masks and also perform various inpainting tasks, such as object insertion, as shown in \cref{fig:qual_insert}.

In \cref{tab:insertion_quantitative}, we evaluate the object insertion task using $CLIP_{dir}$ and a user study. The results indicate that our method also exhibits certain advantages compared to other methods in terms of both semantic alignment and human preference. 
In particular, the total sum of votes in the user study may not be equal to 100\% because a `None of them' option was provided to ensure an unbiased evaluation, allowing participants to reject all candidates if none were satisfactory. 
Detailed information on the setup of the user study, including the specific questions and the number of participants, is provided in the supplementary material.

% \begin{table*}[t]
% \centering
% \caption{\textbf{Quantitative results on the IMFine dataset \cite{shi2025imfine}.} m-LPIPS \cite{zhang2018unreasonable}, m-PSNR and m-FID \cite{heusel2017gans} represent the LPIPS, PSNR and FID scores within the ground truth segmentation masks. Bold numbers indicate the best performance, and underlined numbers represent the second-best results.}
% \begin{adjustbox}{scale=0.9}
% \begin{tabular}{@{}c|cccccc@{}}
% \toprule
%         & m-LPIPS ($\downarrow$) & LPIPS ($\downarrow$) & m-PSNR ($\uparrow$) & PSNR ($\uparrow$) & m-FID ($\downarrow$) & FID ($\downarrow$)         \\ \midrule
% 3DGIC  \cite{huang20253d}       & 0.0356  & 0.3386         & 28.17                      & 20.29        & 205.85 & 200.99   \\
% GScream \cite{wang2024learning} & 0.0296  & 0.2000         & 30.39           & 22.68         & 153.80 & 112.91        \\
% IMFine \cite{shi2025imfine}     & \textbf{0.0177}  & \underline{0.1747} & \underline{33.05} & \underline{23.59}          & \textbf{75.84}  & \underline{57.53}          \\ 
% Ours                            & \underline{0.0195}  & \textbf{0.1685} & \textbf{33.06} & \textbf{23.88} & \underline{83.98}  & \textbf{53.37} \\ \bottomrule
% \end{tabular}
% \end{adjustbox}

% \label{tab:imfine_quantitative}
% \end{table*}

\begin{table*}[t]
\centering
\caption{\textbf{Quantitative results on the IMFine dataset \cite{shi2025imfine}.} 
%LPIPS \cite{zhang2018unreasonable}, PSNR and FID \cite{heusel2017gans} represent the LPIPS, PSNR and FID scores within the ground truth segmentation masks. 
Bold numbers indicate the best performance, and underlined numbers represent the second-best results.}
% \begin{adjustbox}{scale=0.9}
\begin{tabular}{@{}c|ccc@{}}
\toprule
                                & \: LPIPS ($\downarrow$)\:  &\:  PSNR ($\uparrow$)\:  &\:  FID ($\downarrow$)   \:       \\ \midrule
3DGIC  \cite{huang20253d}        & 0.3386                              & 20.29  & 200.99   \\
GScream \cite{wang2024learning} & 0.2000                  & 22.68         & 112.91        \\
IMFine \cite{shi2025imfine}       & \underline{0.1747}  & \underline{23.59}           & \underline{57.53}          \\ 
Ours                              & \textbf{0.1685} & \textbf{23.88}  & \textbf{53.37} \\ \bottomrule
\end{tabular}
% \end{adjustbox}

\label{tab:imfine_quantitative}
\end{table*}

% \begin{table}[]
% \begin{tabular}{@{}c|ccc@{}}
% \toprule
%                             & Gaussian Editor & Infusion & Ours   \\ \midrule
% CLIP Text Image Directional Similarity ($\uparrow$) & 0.0222 & 0.1589   & \textbf{0.1628} \\ \midrule
% User Study    (\%)  &                 &          &       
% \end{tabular}
% \caption{\textbf{Quantitative results of Insertion.} Compare with Bold numbers indicate the best performance.}
% \label{tab:insertion_quantitative}
% \end{table}

\begin{table*}[t]
\centering
\caption{\textbf{Quantitative evaluation of object insertion.} We compare our method against baseline models using the $CLIP_{dir}$ and a user study across four dimensions. Values in the user study columns represent the percentage of user preference votes. Bold numbers indicate the best performance. $CLIP_{dir}$ : CLIP Text-Image Direction Similarity; Align with T.P. : Align with Text Prompt}
\begin{adjustbox}{scale=0.85}
\begin{tabular}{@{}c|ccccc@{}}
\toprule
                & $CLIP_{dir} \uparrow$ & Consistency (\%) & Visual (\%) & Align with T.P. (\%)  & w/o Artifacts (\%) \\ \midrule
Gaussian Editor \cite{chen2024gaussianeditor} & 0.0222  & 20.825 & 23.625  & 13.900 & 17.000               \\
Infusion \cite{liu2024infusion}  & 0.1589  & 31.950  & 31.950     & 26.975  & 30.550                \\ 
Ours            & \textbf{0.1628}  & \textbf{47.225}  & \textbf{44.425}  & \textbf{59.125} & \textbf{45.400}                  \\ \bottomrule
\end{tabular}
\end{adjustbox}
\label{tab:insertion_quantitative}
\end{table*}

\subsection{Qualitative Results}
In \cref{fig:qual_spin}, we present comparisons on the SPIn-NeRF dataset \cite{mirzaei2023spin} using both segmentation and bounding-box masks. The first row shows the input image and its corresponding mask used for training. In the second row, we observe that 3DGIC \cite{huang20253d} fails to preserve the non-inpainted regions and generates white artifacts inside the inpainted regions. 
For GScream \cite{wang2024learning}, the generated content does not align with nearby regions (e.g., the wooden bench in the third and fourth columns). In contrast, our method generates high-fidelity images with multi-view consistency while preserving the non-inpainted regions. Moreover, the inpainting quality is maintained when changing from the segmentation to the bounding-box settings, indicating that our approach is robust to mask shapes.

\begin{figure}[]
    \centering
    \includegraphics[width=\linewidth]{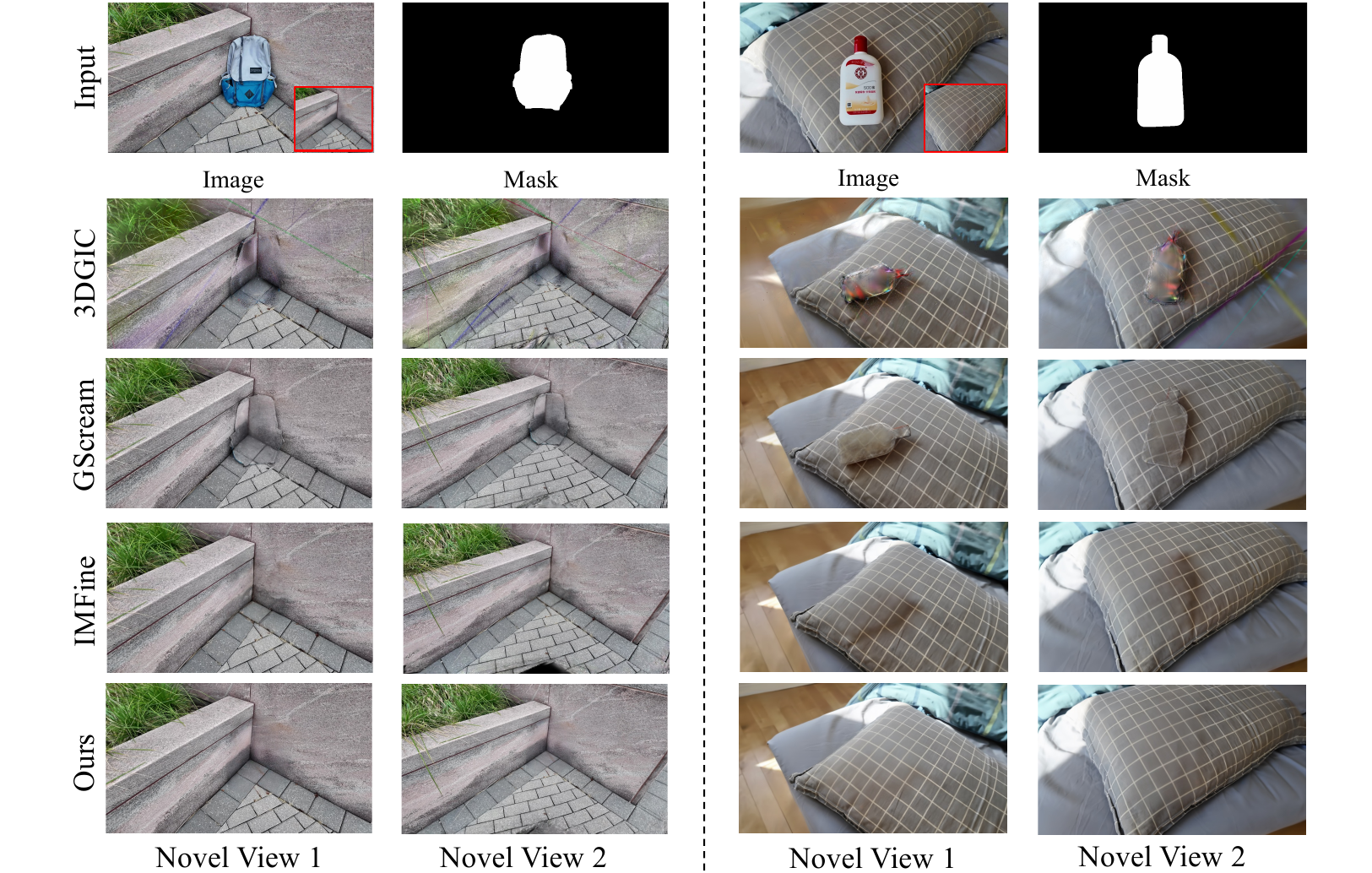}
    \caption{\textbf{Qualitative results on the IMFine dataset \cite{shi2025imfine}}. Same as in \cref{fig:qual_spin}, the first row shows the input images and corresponding masks. We compare the rendering results with 3DGIC \cite{huang20253d}, GScream \cite{wang2024learning}, IMFine \cite{shi2025imfine}. Red box in the input image indicates the reference image for 3DGIC, GScream and Ours.}
    \label{fig:qual_imfine}
\end{figure}

We further compare our method against 3DGIC \cite{huang20253d}, GScream \cite{wang2024learning}, and IMFine \cite{shi2025imfine} on the IMFine dataset, as shown in \cref{fig:qual_imfine}. 3DGIC fails to inpaint removed regions when the geometry is complex, and GScream shows inconsistency when the viewpoint variation is large. Especially in second scene, while IMFine produces high-fidelity images, residual object traces remain after inpainting.

% Beyond removal, we also perform object insertion across several scenes in both datasets. As shown in \cref{fig:qual_insert}, we insert a book with a blue cover into the \textit{``Book"} scene using a segmentation mask, and we insert a melon into the \textit{``Jansport"} scene and a yellow flower into the \textit{``Bucket"} scene, both using bounding box masks. The inserted objects are well aligned with the scene, preserve the surrounding background, and remain consistent across two novel views. Moreover, we present depth maps for each scene to show that the inserted object is well integrated into the 3D scene, not only in the RGB rendering but also in the geometry.

\begin{figure}[]
    \centering
    \includegraphics[width=\linewidth]{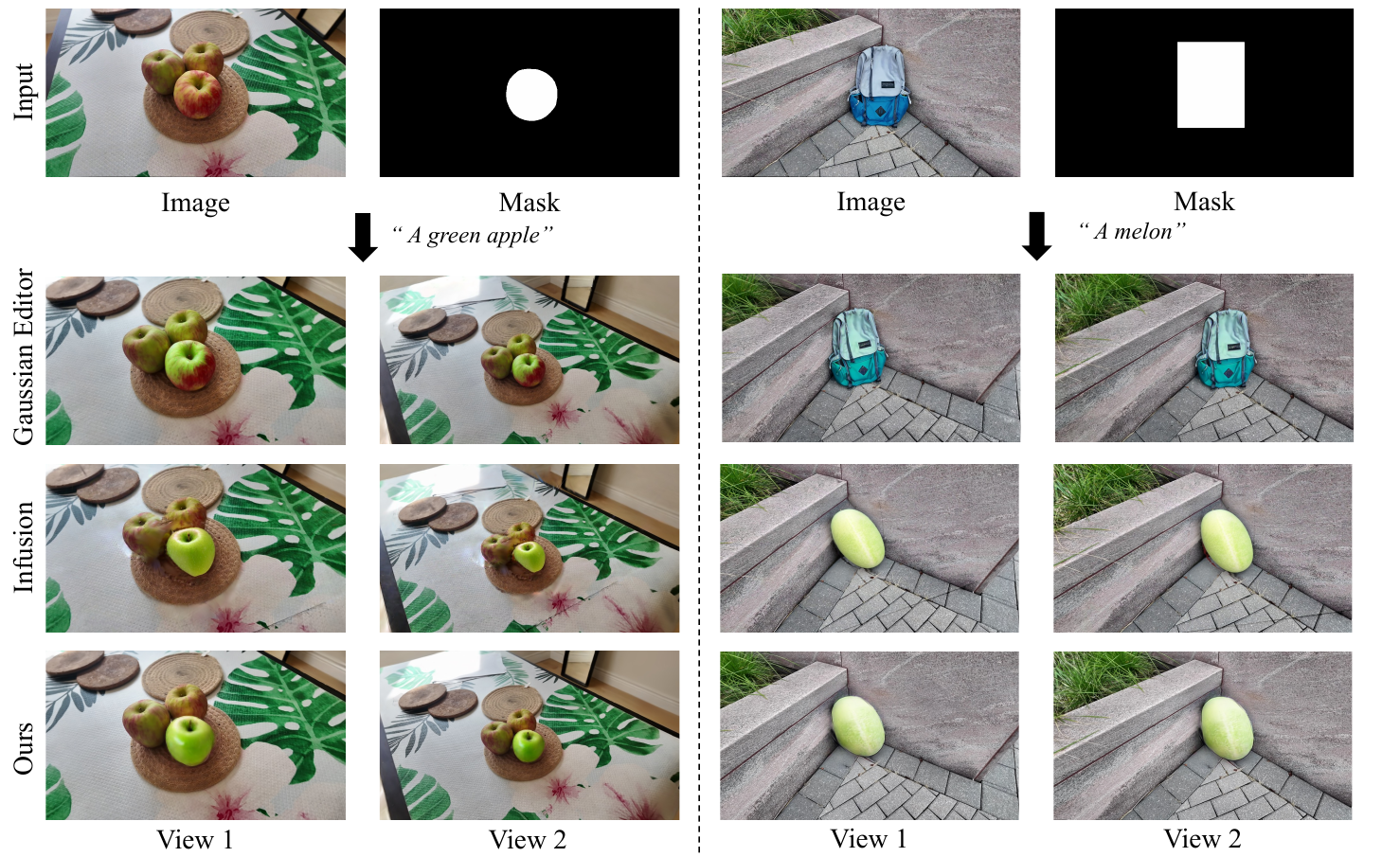}
    \caption{\textbf{Qualitative results for object insertion task.} We perform object insertion tasks on the \textit{``Apples2"} and \textit{``Jansport"} scenes of IMFine dataset \cite{shi2025imfine}. We compare the rendering results with Gaussian Editor \cite{chen2024gaussianeditor} and Infusion \cite{liu2024infusion}.}
    \label{fig:qual_insert}
\end{figure}

Beyond removal, we also perform object insertion across several scenes. As shown in \cref{fig:qual_insert}, Gaussian Editor \cite{chen2024gaussianeditor} fails to synthesize new objects, merely altering colors without handling the required geometric changes, especially in the \textit{``Jansport"} scene. Infusion \cite{liu2024infusion} produces a distorted geometry for the apple in the \textit{``Apples2"} scene and leaves artifacts behind the melon in the \textit{``Jansport"} scene. By contrast, our approach synthesizes high-fidelity objects that are seamlessly integrated into the scene while maintaining strict multi-view consistency across varying viewpoints.

\begin{table*}[t]
\centering
\caption{\textbf{Quantitative results for ablation studies on the SPIn-NeRF dataset \cite{mirzaei2023spin}.} The full model with all key components achieves the best performance.}
% \begin{adjustbox}{scale=0.85}
\begin{tabular}{@{}c|cccccc@{}}
\toprule
Methods  ($\downarrow$)  &\: m-LPIPS \:   & \: LPIPS \:  & \:  m-FID\:   &\:  FID \: \\ \midrule
w/o Ref-GS & 0.0359 & 0.2534 &  87.05  & 30.47 \\
w/o Adaptive Weight  & 0.0356  & 0.2571 & 82.07 & 30.76 \\
w/o Feature Attention & 0.0355 & 0.2559 & 88.15 & 31.78 \\
w/o CLG              & 0.0428 & 0.3057 &  97.04 &  36.47 \\
w/o TE-D             & 0.0535 & 0.2902 &  132.55 & 39.31 \\
Full model (Ours)  & \textbf{0.0317} &\textbf{0.2349} & \textbf{80.44}  & \textbf{28.92} \\ \bottomrule
\end{tabular}
% \end{adjustbox}
% \vspace{-0.5cm}
\label{tab:ablation}
\end{table*}

\subsection{Ablation Study}
\paragraph{Effectiveness of key components.} We conduct ablation studies on the SPIn-NeRF dataset \cite{mirzaei2023spin} to evaluate the effectiveness of our three key components: Reference Adaptive Gaussian Splatting with Feature Attention (Ref-GS), Consistency Loss Guidance (CLG), and Texture-Enhancing Discriminator (TE-D). As shown in \cref{tab:ablation}, removing any component leads to a noticeable decrease in performance in all metrics, indicating that each module is essential for high-quality, consistent inpainting. Specifically, we further examine the individual components of Ref-GS by separately ablating Adaptive Weight and Feature Attention. Excluding either component leads to performance degradation, confirming that both components are vital for aligning the 3DGS scene toward the reference view while maintaining overall structure.
Specifically, removing TE-D significantly decreases accuracies, especially in the m-FID/FID \cite{heusel2017gans}, which indicates that TE-D improves rendering quality with increasing the photometric reality.

\begin{figure}[]
    % \vspace{-0.5cm}
    \centering
    \includegraphics[width=\linewidth]{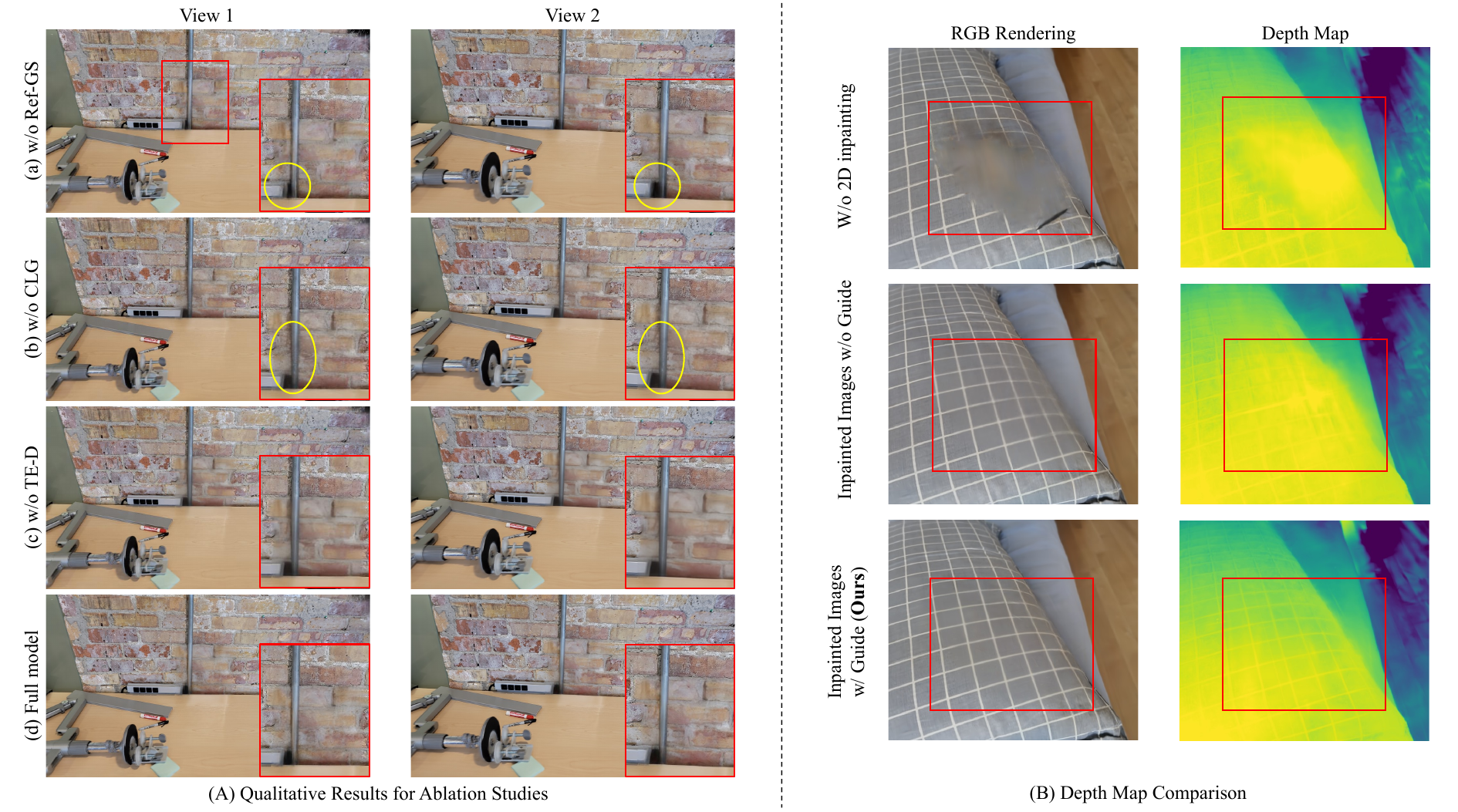}
    \caption{\textbf{Results of ablation studies} (A) Qualitative results for ablation studies on the \textit{``Book"} scene of SPIn-NeRF dataset \cite{mirzaei2023spin}. We mark inpainted regions with red boxes and especially highlight inconsistencies with yellow circles. Rows (a)–(d) show w/o Ref-GS, w/o CLG, w/o TE-D, and the full model, respectively. (B) Depth map comparison on the \textit{``Dabao"} scene from the IMFine dataset \cite{shi2025imfine}. We highlight the inpainted regions with red boxes. For each row, we present the RGB rendering and the depth map from the same viewpoint.}
    \label{fig:full_abl}

\end{figure}

\cref{fig:full_abl}A shows the qualitative comparisons of the ablation study. Rows (a)–(d) correspond to without Ref-GS, without CLG, without TE-D, and the Full model, respectively. Column (a) fails to generate cross-view consistent results (see the yellow circle), and column (b) shows irregular artifacts. Although column (c) achieves consistent results, it produces overly smooth textures compared to the full model. This over-smoothing behavior corresponds to the lowest quantitative scores on all metrics except LPIPS \cite{zhang2018unreasonable}. For LPIPS, the jittering textures observed in the results without CLG lead to even worse scores.

\paragraph{Effectiveness of consistent 2D images on 3D geometry.}
To assess the effect of enforcing 2D consistency during inpainting on the recovered 3D geometry, we present RGB renderings and depth maps in \cref{fig:full_abl}B. The figure reports 3DGS results obtained by training with three types of inputs: occluded 2D images, 2D inpainting outputs without guidance, and our guided images. In the first column, the occluded regions remain visible in both the RGB rendering and the depth map. Although training with unguided 2D inpainting outputs appears to fill the occlusions in the RGB renderings, these renderings remain blurry and the depth maps lack geometric stability. In contrast, ours shows more plausible and consistent RGB rendering and depth map. The results indicate that training 3DGS with consistent 2D images improves not only photometric fidelity but also underlying geometric quality. 

\section{Conclusions}
\label{sec:conclusion}
In this paper, we present CoIn, a comprehensive 2D–3D inpainting framework. By combining the strengths of 2D inpainting and 3D inpainting, our method supports arbitrary-shaped masks while maintaining cross-view geometric and appearance consistency. We achieve this by optimizing the GS scene toward a reference view with adaptive weights (Ref-GS) and using it to guide the 2D inpainter via GS-based Reference Feature Warping. With the guided inpainting images and Texture-Enhancing Discriminator, we fine-tune the inpainted GS scene to obtain multi-view consistency and recover high-frequency textures. In the experiments, we achieve state-of-the-art performance in both quantitative and qualitative aspects. Extensive experiments show that CoIn handles multiple inpainting tasks (e.g., removal and insertion) under both segmentation and bounding-box masks, outperforming existing 3D inpainting approaches.

\par\vfill\par
% \clearpage  % TODO FINAL: This \clearpage needs to be removed from both review and camera-ready versions.

\section*{Acknowledgements}
This work was supported by NST grant (CRC 21015, MSIT), IITP grant (RS-2023-00228996, RS-2024-00459749, RS-2025-25443318, RS-2025-25441313, RS-2026-25526850, RS-2026-25522885, MSIT), KOCCA grant (RS-2024-00442308, MCST) and InnoCORE program (N10260110, MSIT).
% Please insert your acknowledgments here.

\bibliographystyle{splncs04}
\bibliography{main}

\clearpage
% \documentclass[runningheads]{llncs}

% \usepackage[review,year=2026,ID=4924]{eccv}

% % ---------------------------------------------------------------
% % Other packages

% % Commonly used abbreviations (\eg, \ie, \etc, \cf, \etal, etc.)
% \usepackage{eccvabbrv}

% % Include other packages here, before hyperref.
% \usepackage{graphicx}
% \usepackage{booktabs}

% \usepackage{multirow}
% \usepackage[table]{xcolor}
% \usepackage{adjustbox}
% \usepackage{makecell}

% \definecolor{rred}{RGB}{245, 152, 153}
% \definecolor{oorange}{RGB}{253, 205, 154}
% \definecolor{yyellow}{RGB}{248,244,140}

% % The "axessiblity" package can be found at: https://ctan.org/pkg/axessibility?lang=en
% \usepackage[accsupp]{axessibility}  % Improves PDF readability for those with disabilities.

% \usepackage{hyperref}

% % Support for ORCID icon
% \usepackage{orcidlink}

% \begin{document}

% ---------------------------------------------------------------

% \title{CoIn: Comprehensive 2D-3D Inpainting with Gaussian Splatting Guidance \\ \vspace{0.3em}
% {\normalfont\large Supplementary Material}}
% \maketitle

\setcounter{page}{1}
\setcounter{section}{0}
\setcounter{figure}{0}
\setcounter{table}{0}
\setcounter{equation}{0}

\renewcommand{\thesection}{\Alph{section}}

% % camera ready 때는 빼도 되지 않을까??
% \section{Video Results}
% We provide additional qualitative evaluations in the submitted video. 
% The demonstration first presents object removal results on the SPIn-NeRF dataset \cite{sup_spin}, followed by removal results on the IMFine dataset \cite{sup_imfine}, and concludes with object insertion results on both datasets.
% : $\textit{CoIn\_supplementary\_video.mp4}$

\section{Dataset Details}
\label{sup:B}

\subsection{SPIn-NeRF Dataset}
\label{sup:B:spin}
For all experiments on the SPIn-NeRF dataset \cite{sup_spin}, we use all 10 scenes provided in the benchmark. Following prior works, we use the resolution of $1008 \times 567$, released with the dataset. The dataset provides calibrated camera parameters in a simple radial distortion model; therefore, we convert them into a pinhole camera model using the COLMAP \cite{sup_colmap} converter for training 3d Gaussian splatting, while maintaining the original intrinsic and extrinsic parameters. Enclosed bounding-box masks are created from segmentation masks provided by the dataset. These bounding-box masks are directly used as the inpainting regions in our method and also baseline methods for fair comparisons across all scenes.

\subsection{IMFine Dataset}
\label{sup:B:imfine}
We split the IMFine dataset \cite{sup_imfine} into two subsets based on the range of viewpoints. According to the authors of IMFine, the 20 scenes are composed of 2 scenes with $90^\circ$ coverage, 4 with $120^\circ$, 4 with $180^\circ$, and 10 with $360^\circ$ coverage. The viewpoint ranges of $180^\circ$ or less help to demonstrate plausibility with respect to the reference view. While we present results on the 10 scenes (except for $360\,^{\circ}$) in the main paper Table 2, we additionally perform experiments on the scenes of $360^\circ$ in \cref{sup:C:360}.\\

\section{Additional Implementation Details}
Regarding the loss weights, we use $\lambda_r \in \{50, 100\}$ depending on the scene, with $\lambda_D = 10.0$ and $\lambda_{gen} = 0.01$. For the diffusion sampling process, the initial inpainting stage is performed with 50 DDIM steps and a `uniform' discretization schedule. Consistency Loss Guidance (CLG) inpainting stage uses 200 steps and a `quad' schedule. Detailed configuration files and source code will be publicly released to facilitate reproducibility.

\section{User Study Protocol}
To evaluate the perceptual performance of our method, we conducted a user study involving 18 participants. The evaluation was performed on 4 distinct scenes—covering both the SPIn-NeRF \cite{sup_spin} and IMFine datasets \cite{sup_imfine}—including the representative examples presented in the main paper.
For each scene, participants were asked to evaluate the results based on the following four dimensions:
\begin{itemize}
    \item Consistency: Which model exhibits the best multi-view consistency for the inserted object?
    \item Overall visual quality: Which model produces the highest visual quality? (Considering artifacts and color naturalness)
    \item Alignment with text prompts: Which result best aligns with the provided text prompt?
    \item Absence of visual artifacts: Which model best preserves the original background without adding any unwanted elements or artifacts? (Compared with the input image)
\end{itemize}

The study included results from Gaussian Editor \cite{sup_GE}, Infusion \cite{sup_infusion}, and our method, along with a `None of them' option. To ensure an unbiased assessment, the display order of the models was randomized for each question.

\section{Additional Experiments}
\label{sup:C}
We conduct additional experiments on the IMFine datasets \cite{sup_imfine} with the bounding box masks and incorporating the 3D-first method into our pipeline.

\subsection{Bounding Box Masks}
\label{sup:C:BBmask}
To demonstrate the robustness of our method to mask shapes, we present the inpainting results using the bounding-box masks. The same scenes as in the main paper Table 2 are used in the experiments.

\paragraph{Quantitative Results.}
\cref{sup:tab:imfine_bb} shows the quantitative results for the bounding-box mask setting on the IMFine dataset \cite{sup_imfine}. Our method with the bounding-box masks achieves performance comparable to that of the segmentation-mask setting, demonstrating its robustness to arbitrary-shaped masks.

\begin{table}[t]
  \centering
  % 첫 번째 테이블 (왼쪽)
  \begin{minipage}{0.48\textwidth}
    \centering
    % \caption{\textbf{Bounding Box experiments on the IMFine dataset \cite{sup_imfine}.} We present experiments under the bounding-box mask setting. BB mask denotes the bounding-box mask derived from the segmentation mask.}
    \caption{\textbf{Bounding Box experiments on the IMFine dataset \cite{sup_imfine}.} We use the same scenes from the main paper, showing robustness with comparable results. Seg Mask and BB Mask denote the segmentation mask and its derived bounding-box mask, respectively.}
    \label{sup:tab:imfine_bb}
    % \begin{adjustbox}{scale=0.95}
    \begin{tabular}{@{}c|cc@{}}
    \toprule
            & \makecell[c]{Ours \\ (Seg Mask)}  & \makecell[c]{Ours \\ (BB Mask)} \\ \midrule
    LPIPS ($\downarrow$) & 0.1685 & 0.1759             \\
    PSNR ($\uparrow$) & 23.88  & 23.67                \\
    FID ($\downarrow$) & 53.37  & 61.38                \\ \bottomrule
    \end{tabular}
    % \end{adjustbox}
  \end{minipage}
  \hfill % 두 테이블 사이의 간격을 자동으로 조절
  % 두 번째 테이블 (오른쪽)
  \begin{minipage}{0.50\textwidth}
    \centering
    % \caption{\textbf{Quantitative Results of fours scenes with 3D-Removal.} The top three results are highlighted in \textcolor{rred}{red}, \textcolor{oorange}{orange}, and \textcolor{yyellow}{yellow}, respectively.}
    \caption{\textbf{Quantitative results on $360^\circ$ scenes from the IMFine dataset \cite{sup_imfine}.} We compare our 2D-first and 3D-first results for 3D object removal. Bold numbers indicate the best performance.}
    \label{sup:tab:imfine_360_4_scenes}
    % \begin{adjustbox}{scale=0.78}
    % \begin{tabular}{@{}c|ccc@{}}
    % \toprule Mask Type & LPIPS ($\downarrow$) & PSNR ($\uparrow$) & FID ($\downarrow$)   \\ \midrule
                         
    % GScream \cite{sup_gscream}  & 0.1821   & 20.65 & 140.02 \\
    % 3DGIC  \cite{sup_3dgic} & 0.2228  & 20.70  & 208.30 \\
    % IMFine  \cite{sup_imfine}       & \cellcolor{rred} 0.1361   & \cellcolor{yyellow} 22.32 & \cellcolor{rred} 44.42  \\ \midrule
    % Ours (2D-First)          & \cellcolor{yyellow} 0.1463 &  \cellcolor{oorange} 22.33  & \cellcolor{yyellow} 53.02  \\
    % Ours (3D-First) & \cellcolor{oorange} 0.1451  & \cellcolor{rred} 22.67 & \cellcolor{oorange} 48.67  \\ \bottomrule
    \begin{tabular}{@{}c|cc@{}}
    \toprule
            & \makecell[c]{Ours \\ (2D-First)}  & \makecell[c]{Ours \\ (3D-First)} \\ \midrule
    LPIPS ($\downarrow$) & 0.1463 & \textbf{0.1451}             \\
    PSNR ($\uparrow$) & 22.33  & \textbf{22.67}                \\
    FID ($\downarrow$) & 53.02 & \textbf{48.67}                \\ \bottomrule
    \end{tabular}
    % \end{adjustbox}
  \end{minipage}
\end{table}

\paragraph{Qualitative Results.}
In \cref{sup:fig:qual_imfine}, the last row shows qualitative results using the bounding-box masks. 3DGIC \cite{sup_3dgic} and GScream \cite{sup_gscream} produce noticeable artifacts around the object regions, as these 3D-first pipelines rely on accurate segmentation masks to remove objects in the 3D scene. These similar artifacts are also observed in the bottom two rows of \cref{sup:fig:qual_spin}. In contrast, our method maintains stable and coherent inpainting results even with the coarse bounding-box masks, highlighting its benefits where exact object masks are hard to obtain.

\subsection{Incorporating 3D-first Pipeline}
\label{sup:C:360}
\begin{figure}[t]
  \centering
  % 첫 번째 이미지 (왼쪽)
  \begin{minipage}{0.48\textwidth}
    \centering
    \includegraphics[width=\textwidth]{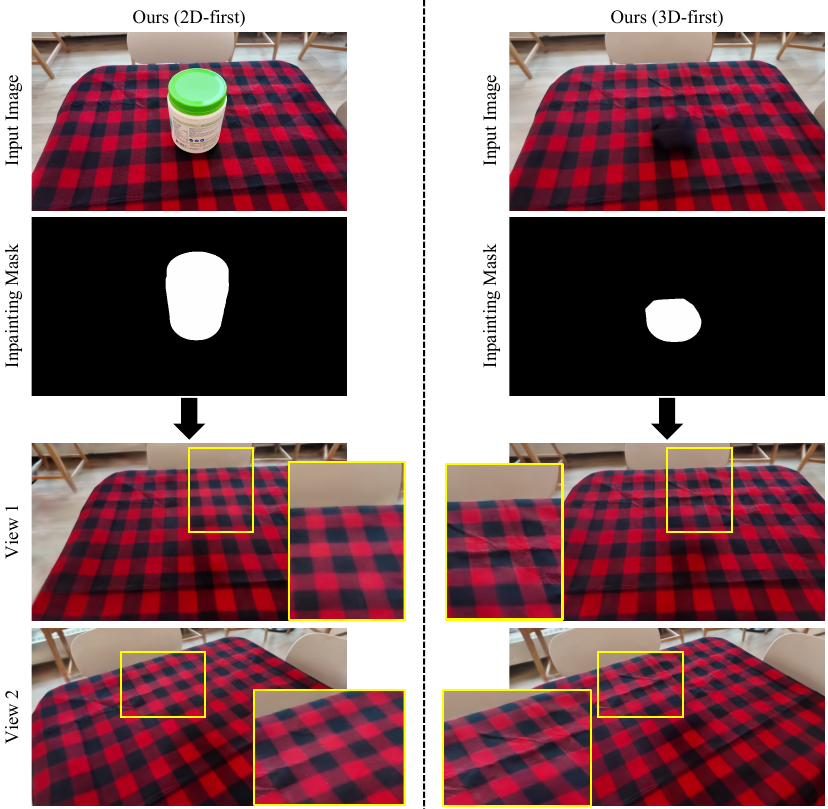}
    \caption{\textbf{Qualitative comparison of 2D-first and 3D-first pipelines.} The 3D-first pipeline simplifies inpainting by targeting only truly occluded areas, unlike the more challenging 2D-first approach with its larger mask.}
    \label{sup:fig:only_ours_360}
  \end{minipage}
  \hfill % 사이 간격 조절
  % 두 번째 이미지 (오른쪽)
  \begin{minipage}{0.48\textwidth}
    \centering
    \includegraphics[width=\textwidth]{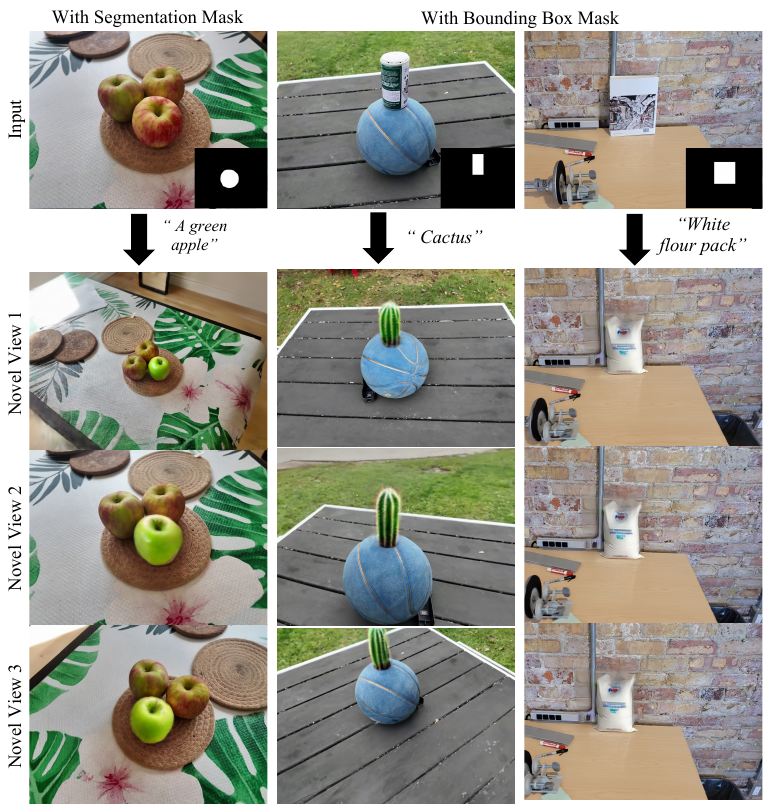}
    % \caption{\textbf{Qualitative result of insertion.} }
    \caption{\textbf{Qualitative results of object insertion.} The first row presents the input images, masks, and the corresponding text prompts.}
    \label{sup:fig:insertion}
  \end{minipage}
\end{figure}

Our method mainly follows a 2D-first pipeline, which shows strength in inpainting with arbitrary-shaped masks and enabling diverse object insertion tasks. However, it receives larger inpainting regions (see the left column of \cref{sup:fig:only_ours_360}), as inpainting is performed on 2D images. Meanwhile, the 3D-first pipeline offers the advantage of inpainting only the truly occluded or missing regions observable across multiple views (see the right column of \cref{sup:fig:only_ours_360}), when certain conditions are met. (i) The mask should be precise to the object in the image, and show effectiveness (ii) when the scene is captured over a wide viewpoint range. Since our method adopts a hybrid 2D–3D design, it can incorporate the benefits of the 3D-first pipeline under certain conditions, allowing us to combine the flexibility of 2D inpainting with the 3D-first pipeline. 

Specifically, we reconstruct a 3DGS scene and remove Gaussian anchors within the masked regions to obtain $\mathcal{G}^{rmv}$. By rendering $\mathcal{G}^{rmv}$ with fixed Gaussian scales, we derive refined 2D masks $M^{rmv}_n$ that represent only the truly occluded regions, along with the corresponding rendered images $R_n^{rmv}$. This process is applied to $360^\circ$ scenes (\textit{``msi", ``desk3", ``rocks", ``bin"}) in the IMFine dataset \cite{sup_imfine}.

For the $360^\circ$ scenes, applying our method directly results in significantly larger inpainting regions compared to the other scenes, making it difficult to achieve competitive performance against 3D-first pipelines such as 3DGIC and IMFine. Therefore, for the object-removal setting in these scenes, we also adopt a mask-based deletion strategy similar to previous 3D-first approaches.

\paragraph{Quantitative Results.}
\cref{sup:tab:imfine_360_4_scenes} shows that incorporating the 3D-first pipeline into our framework leads to a noticeable performance gain in all evaluation metrics. By focusing the inpainting task on a reduced, more precise region, our 3D-first design achieves higher reconstruction quality compared to the 2D-first alternative.

\paragraph{Qualitative Results.}
\cref{sup:fig:only_ours_360} provides a qualitative comparison between the 2D-first and 3D-first pipeline when integrated into our framework. The 3D-first pipeline yields a refined removal mask $M^{rmv}_n$ that is substantially smaller than the one produced by the 2D-first pipeline, as it identifies only the truly occluded regions observed across multiple views. This reduced masked area leads to noticeably more faithful inpainting, resulting in a closer match to the ground-truth images (e.g., wrinkles on the tablecloth). Overall, the qualitative results highlight the benefit of incorporating a 3D-first stage when accurate multi-view cues are available.

\subsection{Robustness to Arbitrary Masks}
As seen in \cref{reb:fig}(a), 2D masks obtained by segmentation model often lack 3D consistency (e.g. including non-target object mask). Our method, however, remains robust to mask qualities, yielding high-quality results even with irregular masks. As shown in \cref{reb:fig}(b), they are obtained by appending random scribbles to object masks. \cref{reb:tab:mask} validates that our method performs comparable to using precise masks, reducing manual efforts, and enhancing utilities.

\begin{table}[]
\centering
\caption{Performance of \textbf{ours across different mask types} on SPin-NeRF dataset \cite{sup_spin} Scene \#10. Best in bold.}
\label{reb:tab:mask}
% \begin{adjustbox}{scale=0.8}
\begin{tabular}{@{}c|cccc@{}}
\toprule
Mask Type  & m-LPIPS ($\downarrow$) & LPIPS ($\downarrow$) & m-FID ($\downarrow$) & FID ($\downarrow$)     \\ \midrule
Irregular mask & 0.0109          & 0.2163          & 64.68          & 19.88   \\
BB mask        & 0.0111          & 0.2179          & \textbf{61.22} & 17.93          \\
Seg mask       & \textbf{0.0106} & \textbf{0.2108} & 65.64          & \textbf{17.28}
\\ \bottomrule
\end{tabular}
% \end{adjustbox}
\end{table}

\subsection{Robustness to Reference View Selection}
As a practical guideline, the reference view is selected based on the desired inpainting result (e.g., the most representative angle of the target object). While this choice is flexible, our method is not sensitive to specific viewpoints; as demonstrated in \cref{reb:tab:ref}, the performance variation across different reference views is small.
\begin{table}[]
\centering
\caption{\textbf{Robustness to reference view selection.} We report the mean and standard deviation across three different reference views (one as used in the main paper and two randomly chosen) on the SPin-NeRF dataset \cite{sup_spin} Scene \#10.}
\label{reb:tab:ref}
% \begin{adjustbox}{scale=0.76}
\begin{tabular}{l|cccc}
\toprule
Metric ($\downarrow$)& m-LPIPS & LPIPS  & m-FID  & FID \\ \midrule
Results & $0.0107 \pm 0.0001$ & $ 0.213 \pm 0.005$ & $60.4 \pm 3.7$ & $18.4 \pm 0.9$ \\ \bottomrule
\end{tabular}
% \end{adjustbox}
\end{table}

\section{Additional Qualitative Results}
\begin{figure*}[t]
    \centering
    \includegraphics[width=\linewidth]{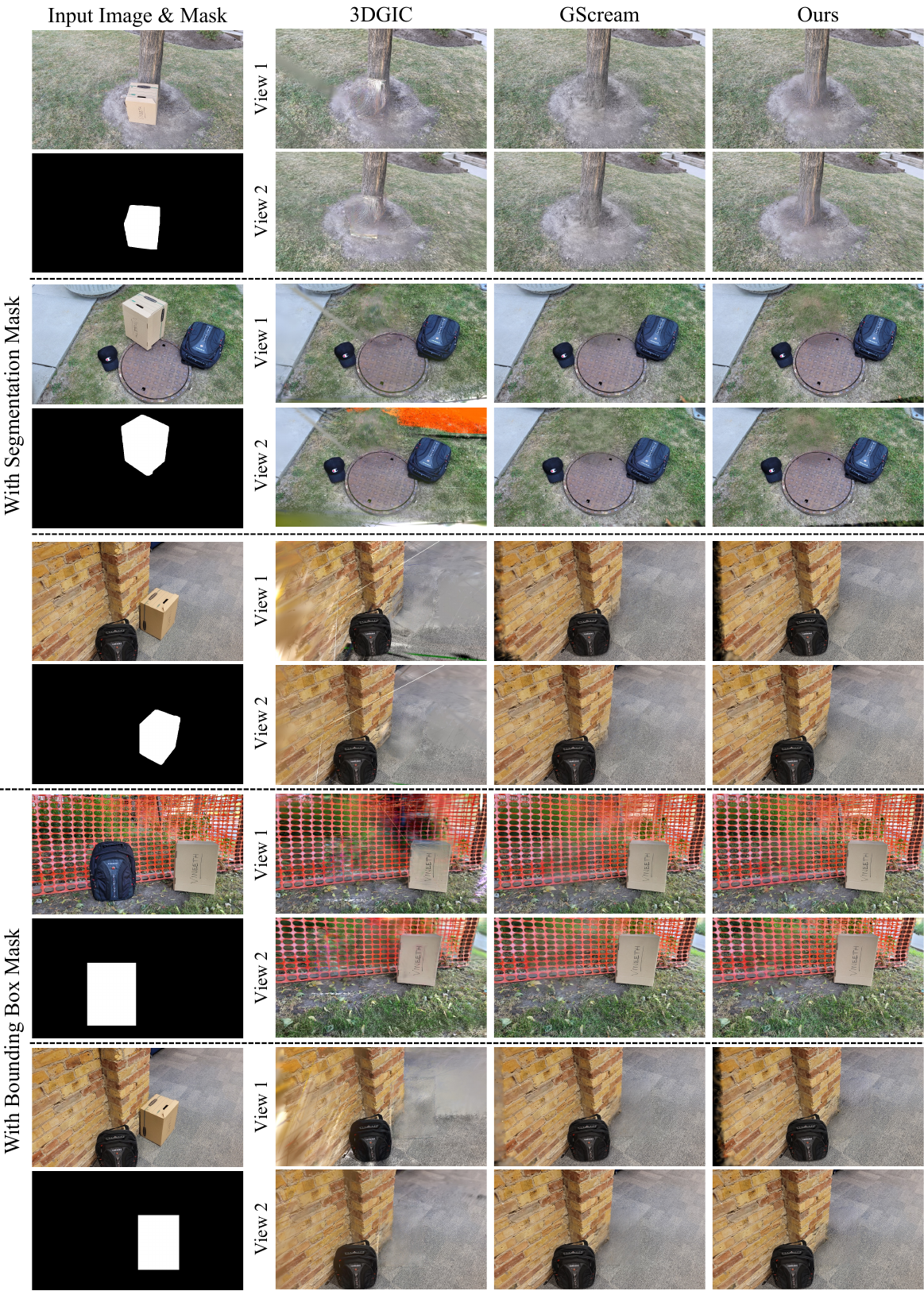}
    \caption{\textbf{Additional qualitative results on the SPIn-NeRF dataset \cite{sup_spin}}. }
    \label{sup:fig:qual_spin}
\end{figure*}

\begin{figure*}[t]
    \centering
    \includegraphics[width=\linewidth]{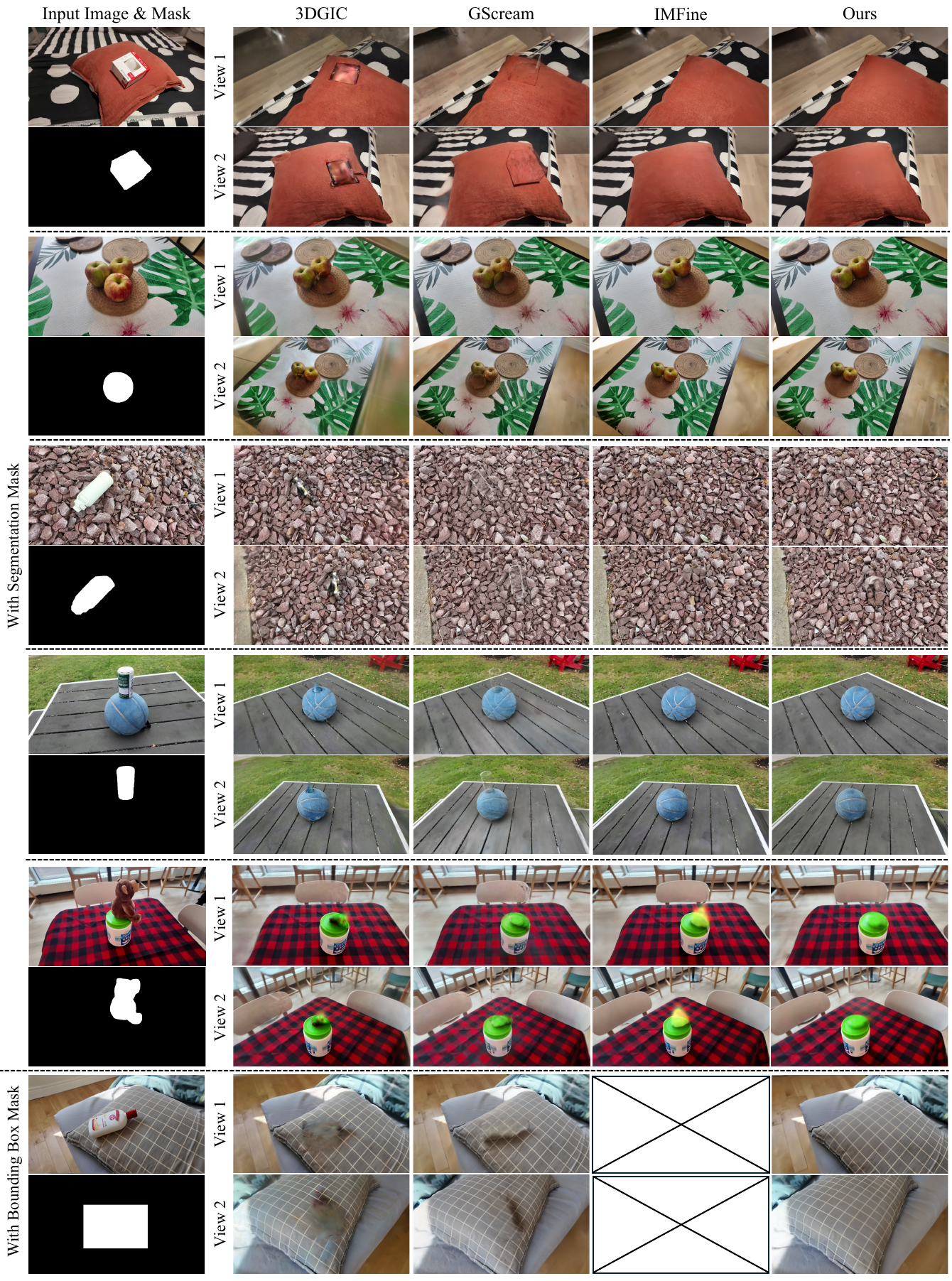}
    \caption{\textbf{Additional qualitative results on the IMFine dataset \cite{sup_imfine}}. }
    \label{sup:fig:qual_imfine}
\end{figure*}

We present several qualitative results on the SPIn-NeRF dataset \cite{sup_spin} in \cref{sup:fig:qual_spin} and IMFine dataset \cite{sup_imfine} in \cref{sup:fig:qual_imfine}. The first column shows the input image and mask. For each paired row, we show different views of the same scene. We present further object insertion experiments in \cref{sup:fig:insertion}, covering a diverse set of scenes from both datasets. These results highlight our method's capability to perform insertion tasks that remain robust across large viewpoint ranges.

\section{Limitation and Future Works}
Our method relies on a 2D inpainting model based on Stable Diffusion (SD) \cite{sup_ldm} 2.0, and therefore the overall performance is constrained by the base model. Although SD-based models accept text prompts, their outputs often depend on the mask shapes, sometimes resulting in reference images that are suboptimal for guiding multi-view consistency. In addition, when the provided mask does not fully cover the target object, the quality of removing the object degrades in the 2D inpainting stage.
Replacing the base inpainting model with a more advanced architecture such as SD-XL \cite{sup_sdxl} could further improve the robustness and visual quality of our pipeline.
\subsection{Failure Analysis}
\begin{figure}[h]
    \centering
    \includegraphics[width=\linewidth]{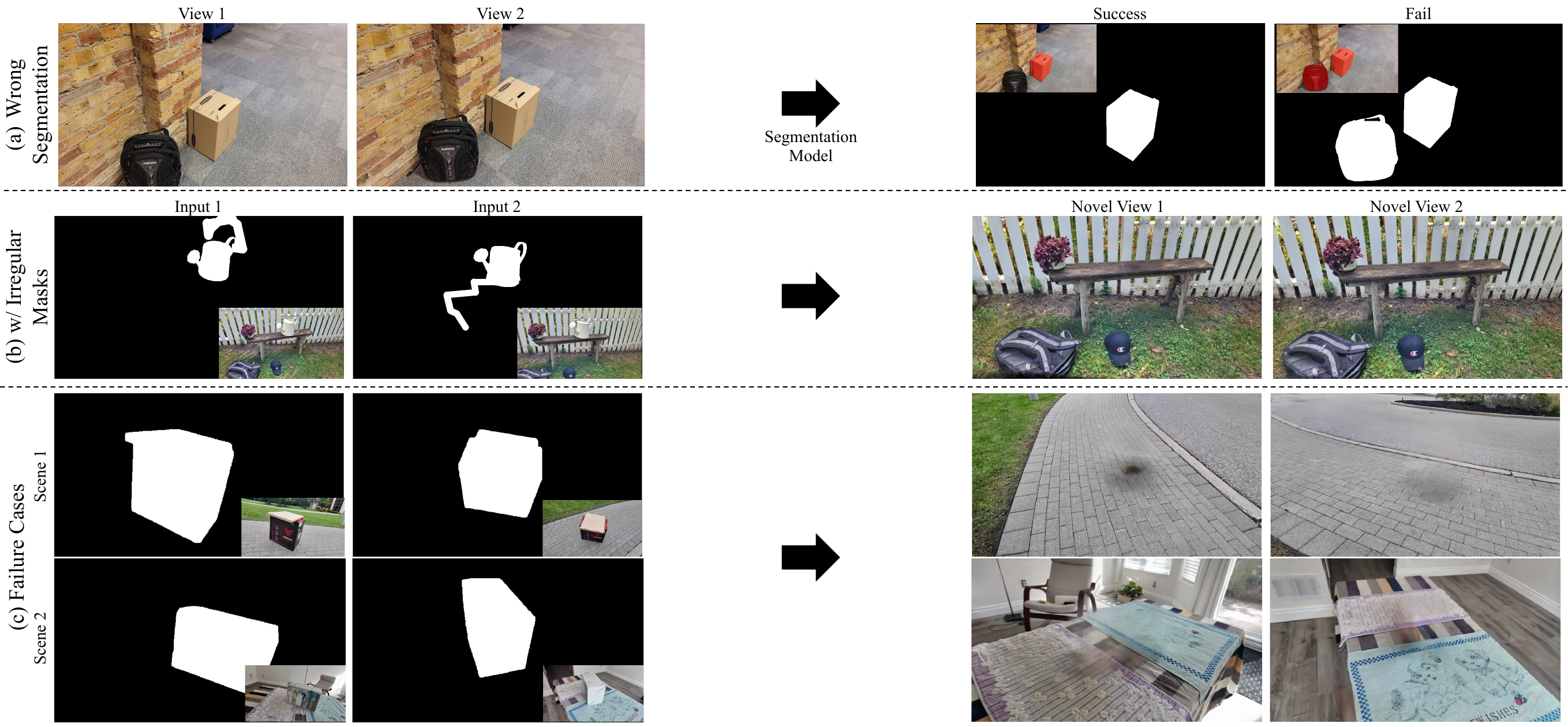}
    \vspace{-0.2cm}
    \caption{\textbf{Qualitative results.} (a) Input views and 3D-inconsistent masks from 2D segmentation. (b) Irregular masks with original inputs (left) and our consistent novel views (right). (c) Failure cases: large masks with view changes beyond $180^\circ$ (left) lead to blurry textures (right).}
    \label{reb:fig}
    \vspace{-0.2cm}
\end{figure}

As shown in \cref{reb:fig}(c), large masks with view changes beyond $180^\circ$ show failure cases due to insufficient context for the 2D inpainting model. However, as discussed in \ref{sup:C:360}, incorporating a 3D-first pipeline addresses this limitation.

\clearpage

% \newpage

% \end{document}
\clearpage
\appendix

\end{document}